\documentclass[conference]{IEEEtran}
\IEEEoverridecommandlockouts

\usepackage{cite}
\usepackage{amsmath}
\usepackage{amssymb,amsfonts}
\usepackage{algorithmic}
\usepackage{graphicx}
\usepackage{textcomp}
\usepackage{xcolor}
\usepackage{adjustbox}
\usepackage{float}
\DeclareUnicodeCharacter{2212}{-}

\def\BibTeX{{\rm B\kern-.05em{\sc i\kern-.025em b}\kern-.08em
    T\kern-.1667em\lower.7ex\hbox{E}\kern-.125emX}}
\begin{document}

\title{Collision-Free Trajectory Planning and control of Robotic Manipulator using Energy-Based Artificial Potential Field (E-APF)\\
\thanks{\textit{*Corresponding author.} \\
\textit{Email address: rakesh.sahoo266@gmail.com (Rakesh Kumar Sahoo)}}
}

\author{\IEEEauthorblockN{Adeetya Uppal}
\IEEEauthorblockA{\textit{Department of Aerospace Engineering} \\
\textit{Indian Institute of Technology}\\
Kharagpur, 721302, India \\
adeetyauppal1@gmail.com}
\and
\IEEEauthorblockN{Rakesh Kumar Sahoo*}
\IEEEauthorblockA{\textit{Department of Aerospace Engineering} \\
\textit{Indian Institute of Technology}\\
Kharagpur, 721302, India \\
rakesh.sahoo266@gmail.com}
\and
\IEEEauthorblockN{Manoranjan Sinha}
\IEEEauthorblockA{\textit{Department of Aerospace Engineering} \\
\textit{Indian Institute of Technology}\\
Kharagpur, 721302, India \\
masinha@aero.iitkgp.ac.in}
 }

\maketitle

\begin{abstract}
Robotic trajectory planning in dynamic and cluttered environments remains a critical challenge, particularly when striving for both time efficiency and motion smoothness under actuation constraints. Traditional path planner, such as Artificial Potential Field (APF), offer computational efficiency but suffer from local minima issue due to position-based potential field functions and oscillatory motion near the obstacles due to Newtonian mechanics. To address this limitation, an Energy-based Artificial Potential Field (APF) framework is proposed in this paper that integrates position and velocity-dependent potential functions. E-APF ensures dynamic adaptability and mitigates local minima, enabling uninterrupted progression toward the goal. The proposed framework integrates E-APF with a hybrid trajectory optimizer that jointly minimizes jerk and execution time under velocity and acceleration constraints, ensuring geometric smoothness and time efficiency. The entire framework is validated in simulation using the 7-degree-of-freedom Kinova Gen3 robotic manipulator. The results demonstrate collision-free, smooth, time-efficient, and oscillation-free trajectory in the presence of obstacles, highlighting the efficacy of the combined trajectory optimization and real-time obstacle avoidance approach. This work lays the foundation for future integration with reactive control strategies and physical hardware deployment in real-world manipulation tasks.
 \end{abstract}


\section{Introduction}
Robotic trajectory planning in dynamic and cluttered environments is a fundamental challenge in autonomous systems. As robots are increasingly deployed in unstructured environments for applications such as assistive manipulation, warehouse automation, and collaborative tasks, the need for generating smooth, dynamically feasible, and time-efficient trajectories becomes crucial. These trajectories must comply with actuation constraints, ensuring real-time performance while navigating around obstacles. Traditional approaches \cite{b1} often struggle to guarantee real-time adaptability and collision-free trajectory without sacrificing motion smoothness or computational efficiency \cite{b2}.  These challenges highlight the need for a path planning framework that ensures robust, safe, and optimal performance in a dynamic environment \cite{b3}. 

The Artificial Potential Field (APF) method is a widely favoured path planning approach of robot due to its computational efficiency and real-time trajectory adjustment in the presence of static and moving obstacles. The fundamental principle involves modelling an artificial landscape where the goal exerts an attractive force and obstacles generate repulsive forces, guiding the robot toward the target while avoiding collisions \cite{b4}. However, traditional APF formulations, which rely solely on position-based potential functions, suffer from local minima trap and oscillatory behaviour near obstacles \cite{b5}. To address these challenges, numerous modifications of APF method have been proposed in the literature. These include redefining the potential functions such as employing superquadric potential field \cite{b6} and rotational potential field \cite{b7} to reduce or eliminate local minima. Alternatively, search based techniques like simulated annealing \cite{b8} and potential gradient descent method \cite{b9} have been explored in literature to escape from local minima traps. Despite these modifications improve the robustness of traditional APF path planners, they introduce computational complexity and may require intricate tuning which limit their practical implementation.

In this paper we have introduced an Energy-based Artificial Potential Field (E-APF) framework that integrates both position and velocity-dependent potential functions to address the inherent limitations of traditional APF. By formulating the attractive and repulsive potential fields in terms of total mechanical energy, the E-APF ensures dynamic adaptability and effectively mitigates the local minima and oscillatory issues commonly observed in position-dependent potential field. To ensure smooth and time-optimal motion, the E-APF is coupled with a hybrid trajectory optimization framework that jointly minimizes jerk and execution time while satisfying velocity and acceleration constraints. The proposed framework is implemented and validated on a 7-degree-of-freedom Kinova Gen3 robotic manipulator in the presence of 3 obstacles. 

The remainder of the paper is structured as follows. Section II presents an overview of the kinematic and dynamic model of the Kinova Gen3 Robotic manipulator. Section III introduces the Energy-based APF and the time-optimal and minimum jerk trajectory optimization is detailed in Section IV. Section V describes the implementation of a Computed torque control strategy for accurate trajectory tracking. Finally, simulation results are presented in Section VI highlighting the practical viability of the method.

\section{Kinematic and Dynamic Modeling of the Kinova Gen3 Robotic Manipulator}

The kinematic model of a robotic system describes the geometric relation between the robot's joint parameters and end-effector's position and orientation. This enables transformations, between the Cartesian space and the joint space, essential for the control and motion planning of a $n$-DOF robotic manipulator, such as the Kinova Gen3 robotic arm. To formulate the kinematic model, a local coordinate frame is assigned to each link of the robotic manipulator using the modified Denavit–Hartenberg (D-H) convention. This convention employs four parameters for each link $i$: $\theta_i$ denotes joint angle, $d_i$ denotes link offset, $\alpha_{i-1}$ denotes link twist, and $a_{i-1}$ denotes link length. The D-H parameters define the spatial relationship between two successive links and are used to construct the homogeneous transformation matrix ${}^{i}_{i-1}T$, which describes the coordinate transformation from frame $i-1$ to frame $i$. This transformation is expressed as a sequence of elementary operations as shown below:
\begin{equation*}
{}^{i}_{i-1}T= \text{R}_{x_{i-1}}(\alpha_{i-1})\times\text{D}_{x_{i-1}}(a_{i-1}) \times \text{R}_{z_{i}}(\theta_i) \times \text{D}_{z_{i}}(d_i)
\end{equation*}
where $\text{R}_{x_{i-1}}$ and $\text{R}_{z_i}$ denote the rotation matrices about the $x$- and $z$-axes, respectively, and $\text{D}_{x_{i-1}}$ and $\text{D}_{z_i}$ represent the translations along the $x$- and $z$-axes, respectively. Each transformation matrix defines the position and orientation of one link relative to the previous one. By recursively multiplying these transformations, the cumulative transformation to the end-effector frame from the base frame can be determined:
\begin{equation*}
{}^0_nT = {}^0_1T \times {}^1_2T \times {}^2_3T \cdots {}^{n-1}_nT
\end{equation*}
The resulting matrix ${}^0_nT$ represents the pose (position and orientation) of the end-effector with respect to the base coordinate frame of the robot. While the kinematic model describes the end-effector's motion based on joint angles without considering the force and torque causing the motion, the dynamic model accounts for the force and torque acting on the system such as inertial effects, Coriolis and centrifugal forces, and gravitational effects. To derive the dynamic equations of motion, the Euler–Lagrange formulation based on energy functions is used as it is suitable for robot system with holonomic constraints. The Euler-Lagrange equation is expressed as,
\begin{equation}
\frac{d}{dt} \left( \frac{\partial \mathcal{L}}{\partial \dot{q}_i} \right) - \frac{\partial \mathcal{L}}{\partial q_i} = 0, \quad \text{for } i = 1, 2, \ldots, n
\end{equation}
Here, $\mathcal{L}$ denotes the Lagrangian of the system, which is defined as the difference between the total kinetic energy $E$ and potential energy $V$ as shown below.
\begin{equation}
\mathcal{L}(q, \dot{q}) = E(q, \dot{q}) - V(q)
\end{equation}
For an n-link robotic manipulator, the total kinetic energy of the system comprises both translational and rotational components associated with each link. The total kinetic energy can be mathematically written as:
\begin{equation}
E = \sum_{i=1}^{n} \left( \frac{1}{2} m_i v_i^T v_i + \frac{1}{2} \omega_i^T \mathbf{I}_i \omega_i \right)
\end{equation}
where $m_i$ denotes the mass of $i$-th link, $v_i$ and $\omega_i$ are its linear and angular velocity vectors, and $\mathbf{I}_i$ represents the inertia matrix of the $i$-link expressed in its local coordinate frame. To map the joint velocities $\dot{q}\in\mathbf{R}^n$ to the cartesian velocities of each link, Jacobian matrices are used. $\mathbf{J}_{v_i}(q)\in\mathbf{R}^{3\times n}$ and $\mathbf{J}_{\omega_i}(q)\in\mathbf{R}^{3\times n}$ denote the translational and rotational Jacobians. Substituting this in the expression of total kinetic energy as:
\begin{equation}
    E=\frac{1}{2}\dot{q}^{T}\left[\sum_{i=1}^{n}\left(m_{i}\boldsymbol{J}_{v_{i}}^{T}\boldsymbol{J}_{v_{i}}+\boldsymbol{J}_{\omega_{i}}^{T}\boldsymbol{R}_{i}\boldsymbol{I}_{i}\boldsymbol{R}_{i}^{T}\boldsymbol{J}_{\omega_{i}}\right)\right]\dot{q}
    \label{kin_energy}
\end{equation}
The expression of total kinetic energy \ref{kin_energy} can be written in compact form as shown below.
\begin{equation}
E = \frac{1}{2} \dot{q}^T M(q) \dot{q}
\end{equation}
where $M(q)\in\mathbf{R}^{n\times n}$ is the configuration-dependent inertia matrix of the manipulator given by
\begin{equation*}
M(q) = \sum_{i=1}^{n} \left( m_i \mathbf{J}_{v_i}^T \mathbf{J}_{v_i} + \mathbf{J}_{\omega_i}^T \mathbf{R}_i \mathbf{I}_i \mathbf{R}_i^T \mathbf{J}_{\omega_i} \right)
\end{equation*}
where $\mathbf{R}_i\in \mathbf{R}^{3\times3}$ is the rotation matrix that transforms from local coordinate frame of $i$-link frame to the base frame.

The potential energy $V$, due to gravitational effect acting on the center of mass of each link is expressed as:
\begin{equation}
V = \sum_{i=1}^{n} m_i \mathbf{g}^T \mathbf{p}_{c_i}
\end{equation}
where $\mathbf{g}$ is the gravitational acceleration vector and $\mathbf{p}_{c_i}\in \mathbf{R}^3$ is the position vector of the center of mass of the $i$-th link in the base frame. By substituting $T$ and $V$ to the Euler–Lagrange equation, the equation of motion of the robotic manipulator can be derived in the standard second-order form as:
\begin{equation}
M(q) \ddot{q} + C(q, \dot{q}) \dot{q} + G(q) = \boldsymbol{\tau}
\label{dynamics}
\end{equation}
Here, $\ddot{q}$ and $\dot{q}$ are the joint acceleration and velocity vectors, $C(q, \dot{q})$ accounts for Coriolis and centrifugal effects, $G(q)$ denotes the gravitational effects, and $\boldsymbol{\tau}$ is the vector of joint control torques applied by actuator.
\,

\section{Artificial Potential Field based Path Planning Algorithm}
In this section, we have introduced a novel Artificial Potential Field-based path planning algorithm which is free from local minima.

\subsection{Traditional Artificial Potential Field}
\begin{figure}[H]
    \centering
    \includegraphics[width=\linewidth]{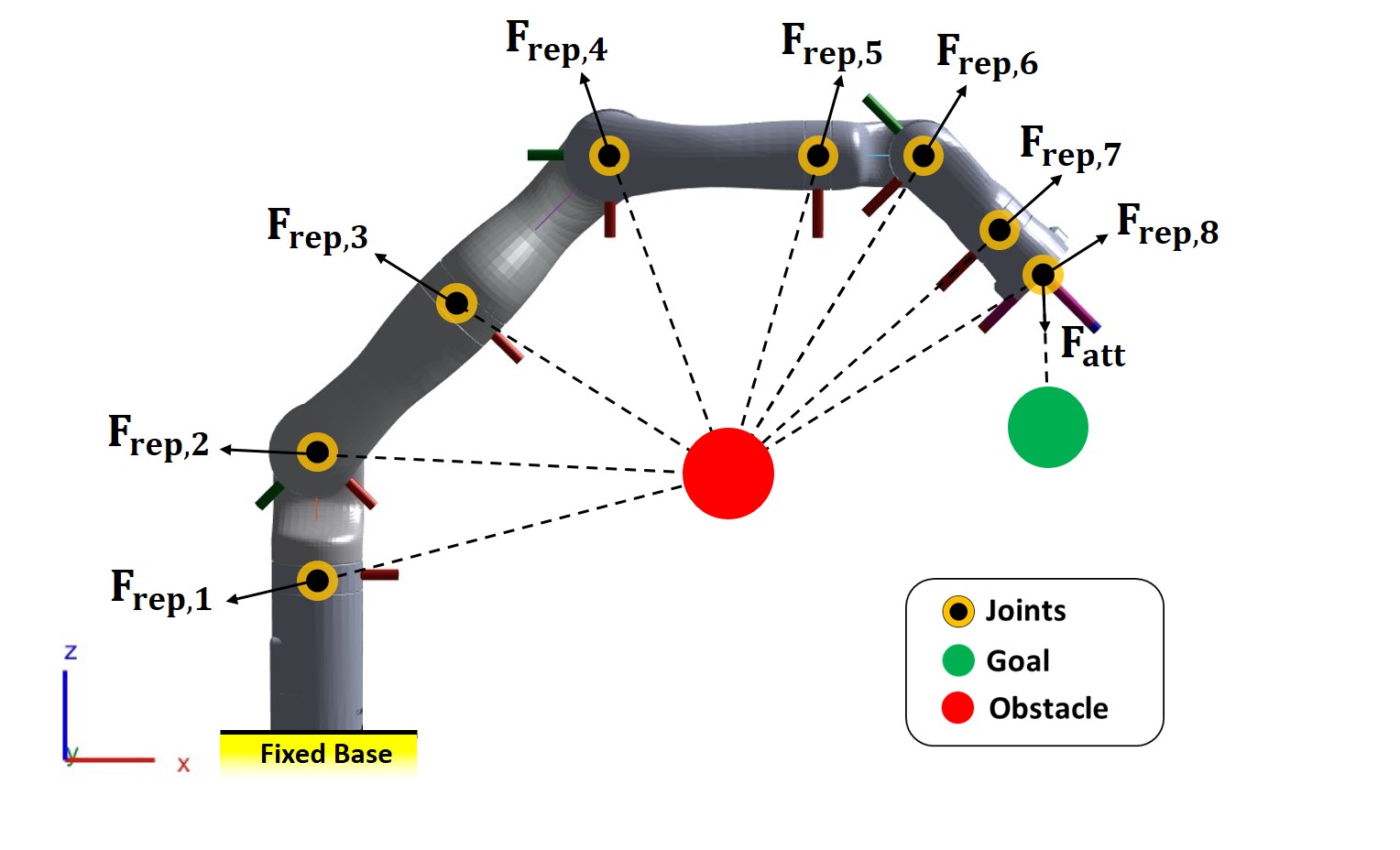}
    \caption{Kinova Gen3 Robotic Manipulator}
    \label{fig:apf}
\end{figure}

The Artificial Potential Field (APF ) algorithm is a widely adopted method for collision-free path planning in autonomous systems, owing to its computational efficiency and ease of implementation. In this framework, the robot and obstacles are modeled as positively charged particles, while the goal is modeled as a negatively charged particle. The attractive force directs the robot toward the goal, whereas the repulsive force from neighbouring obstacles drive it away. The end effector and joints should not collide with obstacles, as shown in Fig. \ref{fig:apf}. The attractive and repulsive potential function described in \ref{apf_func} is defined in terms of robot's relative position w.r.t goal ($r_e$) and w.r.t the obstacle ($r_o$), respectively as shown below.
\begin{equation}
    \begin{array}{ccc}
U_{a}=\frac{1}{2}k_{a}\left(r_{e}\right)^{2} & , & U_{r}=\frac{1}{2}k_{r}\left(\frac{1}{r_{o}}-\frac{1}{\rho_{o}}\right)^{2}\end{array}
\label{apf_func}
\end{equation}
where $\rho_o$ is the radius of the sphere of influence of the repulsive field. The net APF force acting on the robot as shown in \ref{apf_force}  is derived from the gradient of the potential function, and is given as
\begin{equation}
    \begin{aligned}F_{apf} & =-\nabla U_{a}-\nabla U_{r}\\
 & =-k_{a}r_{e}+k_{r}\left(\frac{1}{r_{0}}-\frac{1}{\rho_{0}}\right)\left(\frac{1}{r_{0}}\right)^{2}
\end{aligned}
\label{apf_force}
\end{equation}
It can be inferred from the expression of $F_{apf}$ that when the magnitude of the first and second term becomes equal, net APF force vanishes. This condition corresponds to the local minima problem where the attractive and repulsive force balance out each other and the traditional APF algorithm fails to generate a collision-free path.
\subsection{Energy-based Artificial Potential Field}
To address the limitations of traditional APF, we have proposed an Energy-based Artificial Potential Field (E-APF) framework that incorporates both relative position and velocity into the APF force formulation. Unlike classicial Newtonian-based traditional APF approach, E-APF is modeled using Lagrangian mechanics, enabling smoother trajectory by favoring minimum potential path rather than computing net force.

The attractive static potential $U_a$ and attractive kinetic potential $K_a$, defined as the function of relative position $r_e$ and relative velocity $v_e$, respectively with respect to goal is given by 
\begin{equation}
    \begin{array}{ccc}
U_{a}=\frac{1}{2}k_{a}\left(r_{e}\right)^{2} & , & K_{a}=\frac{1}{2}\gamma k_{a}\left(v_{e}\right)^{2}\end{array}
\label{eapf_att_func}
\end{equation}
where $\gamma \in (0,1)$ denotes the velocity scaling factor. The attractive Lagrangian operator is defined as the difference between attractive kinetic and static potential as shown below.
\begin{equation*}
    \begin{aligned}L_{a}(r_{e},v_{e}) & =K_{a}-U_{a}\\
 & =\frac{1}{2}\gamma k_{a}\left(v_{e}\right)^{2}-\frac{1}{2}k_{a}\left(r_{e}\right)^{2}
\end{aligned}
\end{equation*}
The corresponding attractive force $F_a$ in E-APF is computed by solving the Euler-Lagrangian equation as shown below.
\begin{equation}
    \begin{aligned}F_{a} & =\frac{d}{dt}\left(\frac{\partial L_{a}}{\partial v_{e}}\right)-\frac{\partial L_{a}}{\partial r_{e}}\\
F_{a} & =\gamma k_{a}\dot{v}_{e}-k_{a}r_{e}
\end{aligned}
\label{eapf_att_force}
\end{equation}
Similarly, the repulsive static potential $U_r$  and repulsive kinetic potential $K_r$ is defined as the function of relative position $r_o$ and relative velocty $v_o$, respectively with respect to obstacle is given by 
\begin{equation}
    \begin{array}{ccc}
U_{r}=\frac{1}{2}k_{r}\left(\frac{1}{r_{o}}-\frac{1}{\rho_{o}}\right)^{2} & , & K_{r}=\frac{1}{2}\gamma k_{r}\left(\frac{1}{v_{o}}-\frac{1}{\mu_{o}}\right)^{2}\end{array}
\label{eapf_rep_func}
\end{equation}
where $\rho_o$ and $\mu_o$ denote the spatial and velocity influence bounds of the repulsive field, $\gamma \in (0,1)$ is  velocity scaling factor. The repulsive Lagrangian operator is defined as the difference between repulsive kinetic and static potential as shown below.
\begin{equation*}
    \begin{aligned}L_{r}(r_{o},v_{o}) & =K_{r}-U_{r}\\
 & =\frac{1}{2}\gamma k_{r}\left(\frac{1}{v_{o}}-\frac{1}{\mu_{o}}\right)^{2}-\frac{1}{2}k_{r}\left(\frac{1}{r_{o}}-\frac{1}{\rho_{o}}\right)^{2}
\end{aligned}
\end{equation*}
The repulsive force $F_r$ in E-APF is computed by solving the Euler-Lagrangian equation as shown below.
\begin{equation}
    \begin{aligned}F_{r} & =\frac{d}{dt}\left(\frac{\partial L_{r}}{\partial v_{o}}\right)-\frac{\partial L_{r}}{\partial r_{o}}\\
F_{r} &= \gamma k_{r}\left(\frac{3}{v_{o}}-\frac{2}{\mu_{o}}\right)\frac{\dot{v}_{o}}{v_{o}^{3}}+k_{r}\left(\frac{1}{r_{0}}-\frac{1}{\rho_{0}}\right)\left(\frac{1}{r_{0}}\right)^{2}
\end{aligned}
\label{eapf_att_force}
\end{equation}
By incorporating both relative position and velocity into the potential function formulation, the E-APF approach ensures a more robust and efficient path planning process. Even when the positional attractive and repulsive forces counterbalance, the velocity components ensure continued progression toward the goal, facilitating robust obstacle avoidance and dynamic adaptability. The total E-APF force based on the proposed E-APF approach is expressed as follows:
\begin{equation}
\begin{aligned}F_{e-apf}=\,\, & F_{a}+F_{r}\\
=\,\, & \gamma k_{a}\dot{v}_{e}-k_{a}r_{e}+\gamma k_{r}\left(\frac{3}{v_{o}}-\frac{2}{\mu_{o}}\right)\frac{\dot{v}_{o}}{v_{o}^{3}}+\\
 & k_{r}\left(\frac{1}{r_{0}}-\frac{1}{\rho_{0}}\right)\left(\frac{1}{r_{0}}\right)^{2}
\end{aligned}
\end{equation}
\,

\section{Trajectory Optimization}
The trajectory optimization framework aims to compute a dynamically feasible, smooth, and time-efficient joint space trajectory for a 7-DOF Kinova Gen3 robotic manipulator. The objective function \ref{obj_func} balances motion smoothness and execution speed by minimizing the integrated squared jerk along the trajectory while penalizing the longer execution time. The objective cost function is defined as shown below. 
\begin{equation}
F = \min_{\mathbf{q}(t), T_f} \quad \int_0^{T_f}\underbrace{\left\| \dddot{\mathbf{q}}(t) \right\|^2 dt}_{\text{Minimum Jerk}} \quad+ \underbrace{\lambda T_f}_{\text{Time Penalty}}
\label{obj_func}
\end{equation}
where $q\in\mathbf{R}^7$ represents the joint angle vector and $T_f$ is the total trajectory time. The first term of \ref{obj_func} ensures minimum jerk motion, whereas the second term imposes a time penalty with $\lambda >0$ governing the tradeoff between trajectory smoothness and execution speed. Higher $\lambda$ results in a faster trajectory tracking, while a smaller $\lambda$ prioritizes smoothness. The optimization is subjected to the following joint-level inequality constraints:\\
\\
\textbf{Velocity Constraint}
\begin{equation}
-10 \leq \dot{\mathbf{q}}(t) \leq 10
\end{equation}
This constraint ensures that the joint angular velocities of the robotic manipulator lie within the safe operational bound of ±10 rad/s that could compromise stability or cause damage.\\
\\
\textbf{Acceleration Constraint}
\begin{equation}
-50 \leq \ddot{\mathbf{q}}(t) \leq 50
\end{equation}
This constraint prevents abrupt dynamic torque load on each joint of the robotic manipulator, ensuring that the joint accelerations lies within the range of ±50 rad/s\textsuperscript{2}.

The optimized states of desired joint position $q_d$, angular velocity $\dot{q}_d$, and angular acceleration $\ddot{q}_d$ are subsequently used to compute the instantaneous tracking error. These errors are then fed as input to the computed torque controller to ensure accurate trajectory tracking. 
\,
\section{Computed Torque Controller}

\begin{figure}[H]
    \centering
    \includegraphics[width=\linewidth]{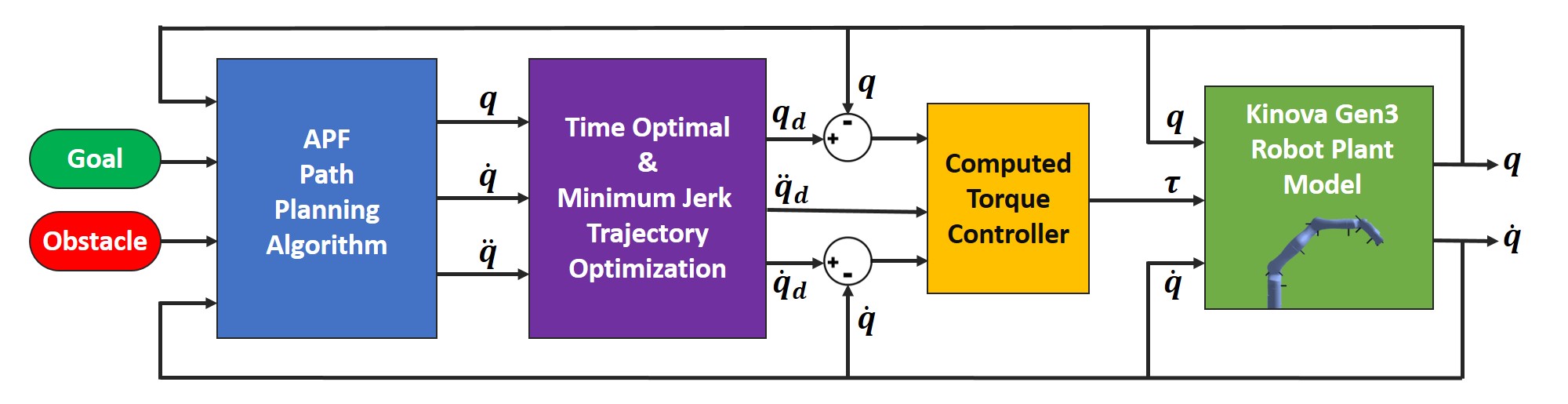}
    \caption{Schematic layout of the proposed approach}
    \label{fig:Layout}
\end{figure}

Computed Torque Control (CTC) is used as a controller to compensate for the non-linear coupled dynamics of robot through feedback linearization approach. This allows the transformation of the complex nonlinear system into a set of decoupled linear second-order subsystems. A Proportional Derivative (PD) control law tracks the joint angle and joint angular velocity errors for each joint independently. For the Kinova Gen3, the joint space dynamics can be described as shown in \ref{dynamics}. $\tau\in \mathbf{R}^7$ is the vector of control torque applied at each joint. The full-state feedback control law is given by
\begin{equation}
\tau = M(q)\left[\ddot{q}_d + k_p*q_e + k_d*\dot{q}_e\right] + C(q, \dot{q})\dot{q} + G(q)
\label{control law}
\end{equation}
where $q_e=q_d - q$ and $\dot{q}_e=\dot{q}_d - \dot{q}$ denotes the relative joint angle and velocity; $q_d$, $\dot{q}_d$, $\ddot{q}_d$ denote the desired joint position, joint velocity, and joint acceleration, respectively;   $k_p, k_d >0$ denote the proportional and derivative gains, respectively of PD control ensuring exponential convergence of the tracking error to zero. Substituting the control law \ref{control law} into the joint space dynamics of the robot \ref{dynamics},  the closed-loop error dynamics is simplified to a set of second-order linear differential equation as shown below.
\begin{equation*}
\ddot{\tilde{q}}_d + k_d*\dot{q}_e + k_p*q_e = 0
\end{equation*}
The overall control architecture integrates motion planning and control in a modular structure as shown in Fig. \ref{fig:Layout}. Initially, a collision-free trajectory is generated by the Artificial Potential Field (APF) path planning algorithm by accounting for goal attraction and obstacle avoidance. The resulting reference positions and velocities are then refined using a time-optimal and minimum jerk trajectory optimization module to ensure smooth and dynamically feasible motion. These optimized trajectories serve as inputs to the Computed Torque Controller (CTC), which compensates for the nonlinear dynamics of the Kinova Gen3 manipulator and generates joint torques accordingly. The plant model executes the motion and provides feedback for closed-loop control. 

\section{Results and Discussion}
The proposed Energy-based Artificial Potential Field (E-APF) approach was comprehensively evaluated against the traditional Artificial Potential Field (APF) method using the Kinova Gen3 Robotic Manipulator. The start configuration of the robotic manipulator was the home configuration $[0, 0, 0, 0, 0, 0, 0]$ while the goal configuration was given as $[1.5, −0.8, 0.5, −0.9, 0.6, 0.3, −0.4]$ to test the robotic manipulator's ability to reach a non-trivial goal configuration. 

Three obstacles were defined in the environment that the robot manipulator had to avoid and maneuver accordingly. These obstacles included a cylindrical collision object with a height of 0.15m and radius of 0.075m placed at the coordinates [0, 0.7, 0.90] m, and two spherical collision objects each with a radius of 0.1m placed at the coordinates [0, 0.3, 1.0] m and [-0.3, 0.1, 1.1] m in the 3D environment. The parameters of E-APF and APF approaches, such as $k_{\text{a}}$, $k_{r}$, $\rho_0$, and $\gamma$ are given in Table \ref{apf_params_table}. The velocity influence bounds $\mu_0$ were dynamically adjusted based on the proximity of the obstacle and an enhanced force balancing was applied by utilizing the position and velocity components.
\begin{figure}[H]
    \centering
    \includegraphics[width=\linewidth]{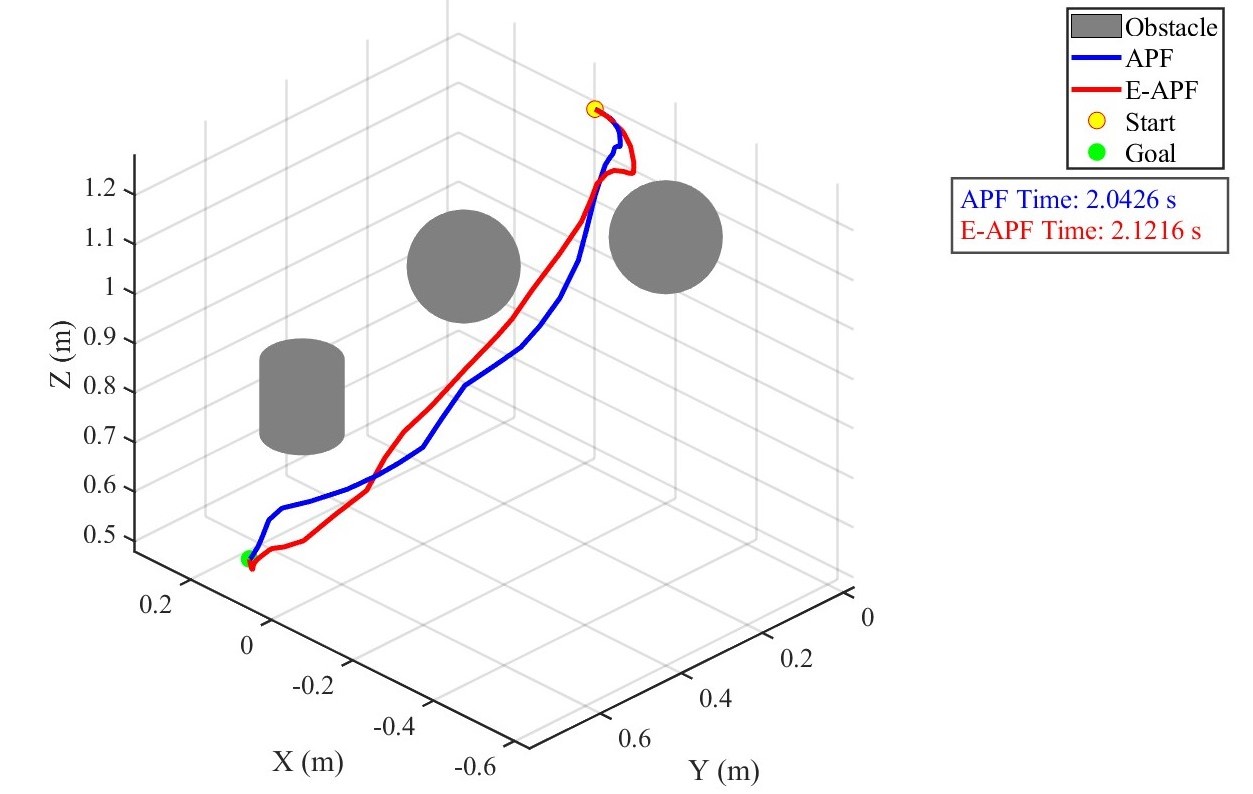}
    \caption{3D Trajectory Comparison between APF and E-APF Methods}
    \label{fig:3d_trajcomp}
\end{figure}
\begin{table}[h]
    \centering
    \caption{APF and E-APF Algorithm Parameters}
    \begin{tabular}{@{} l l l @{}}
        \hline
        \textbf{Parameter} & \textbf{APF} & \textbf{E-APF} \\
        \hline
        Attractive Gain ($k_{\text{att}}$) & 5.0 & 5.0 \\
        Repulsive Gain ($\eta_{\text{rep}}$) & 10.0 & 10.0 \\
        Obstacle Influence Distance ($\rho_0$) & 0.4 m & 0.4 m \\
        Velocity Scaling Factor ($\gamma$) & - & 0.8 \\
        \hline
    \end{tabular}
    \label{apf_params_table}
\end{table}
For the traditional APF controller, the proportional and derivative gain were considered as 25 and 10, respectively, whereas for E-APF gains $k_p$ and $k_d$ were chosen as 49 and 11.2, respectively.

The 3D trajectory plot in Fig. \ref{fig:3d_trajcomp} compares the collision-free path planned by Energy-based APF (red) and the traditional APF (blue) in presence of three obstacles. The traditional APF trajectory exhibits multiple sharp turns and oscillatory behaviour near the obstacles due to the competing nature of attractive and repulsive forces, a known limitation of the position-only dependent potential function of traditional APF. In contrast, the E-APF trajectory is noticeably smoother, maintains consistent directional flow as it incorporates both position and velocity-dependent potential functions. This dual-dependence allows momentum-aware trajectory corrections while mitigating local minima, demonstrating the effectiveness of the velocity-dependent terms in dampening oscillatory motion and maintains a safer distance from the obstacles. Although the APF reaches the goal in 2.0426 s and E-APF in 2.1216 s, the marginal increase in time with E-APF is due to the improved trajectory smoothness.

\begin{figure}[H]
    \centering
    \includegraphics[width=\linewidth]{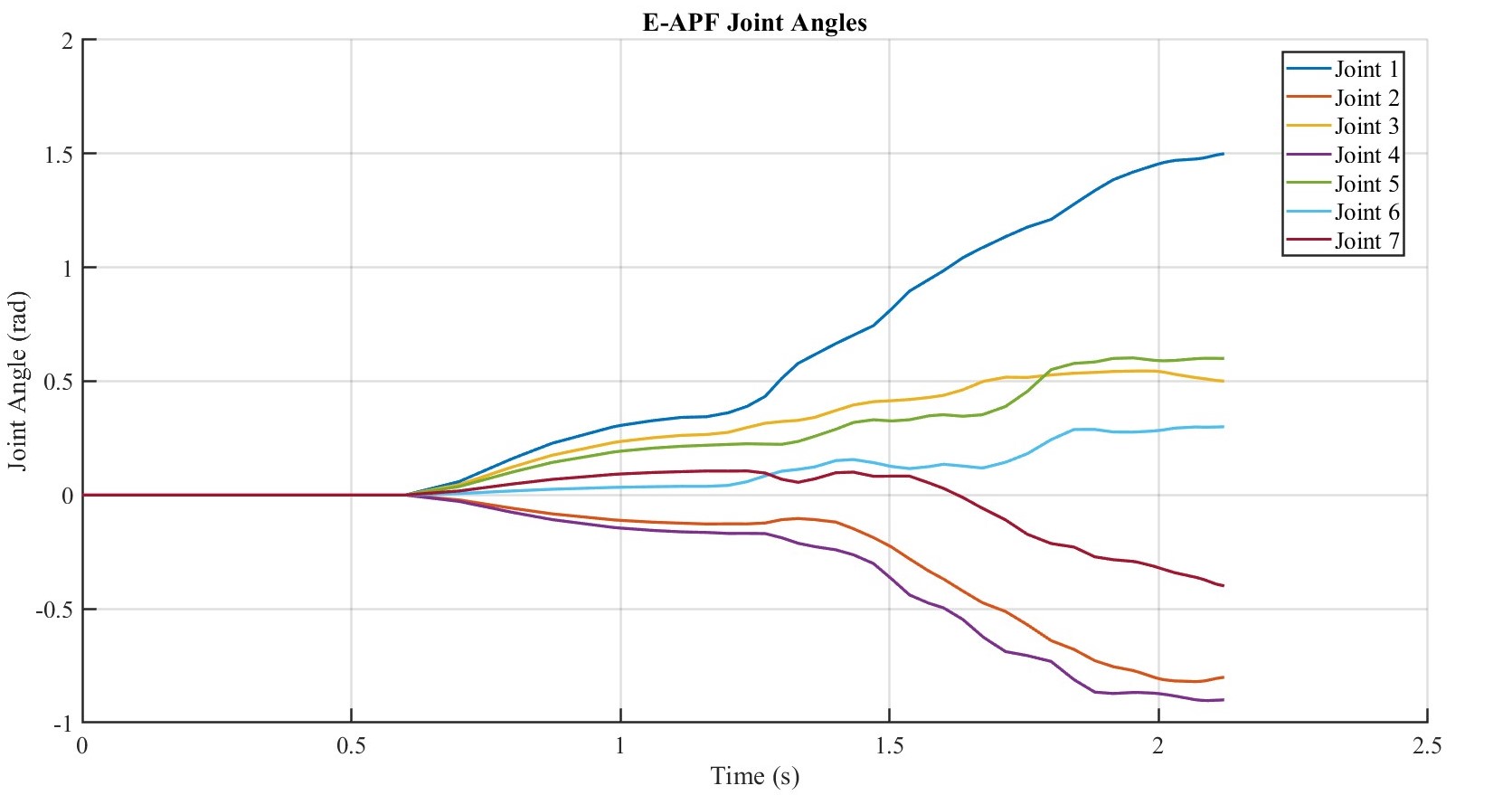}
    \caption{E-APF Joint Angle Evolution}
    \label{fig:eapf_angles}
\end{figure}

\begin{figure}[H]
    \centering
    \includegraphics[width=\linewidth]{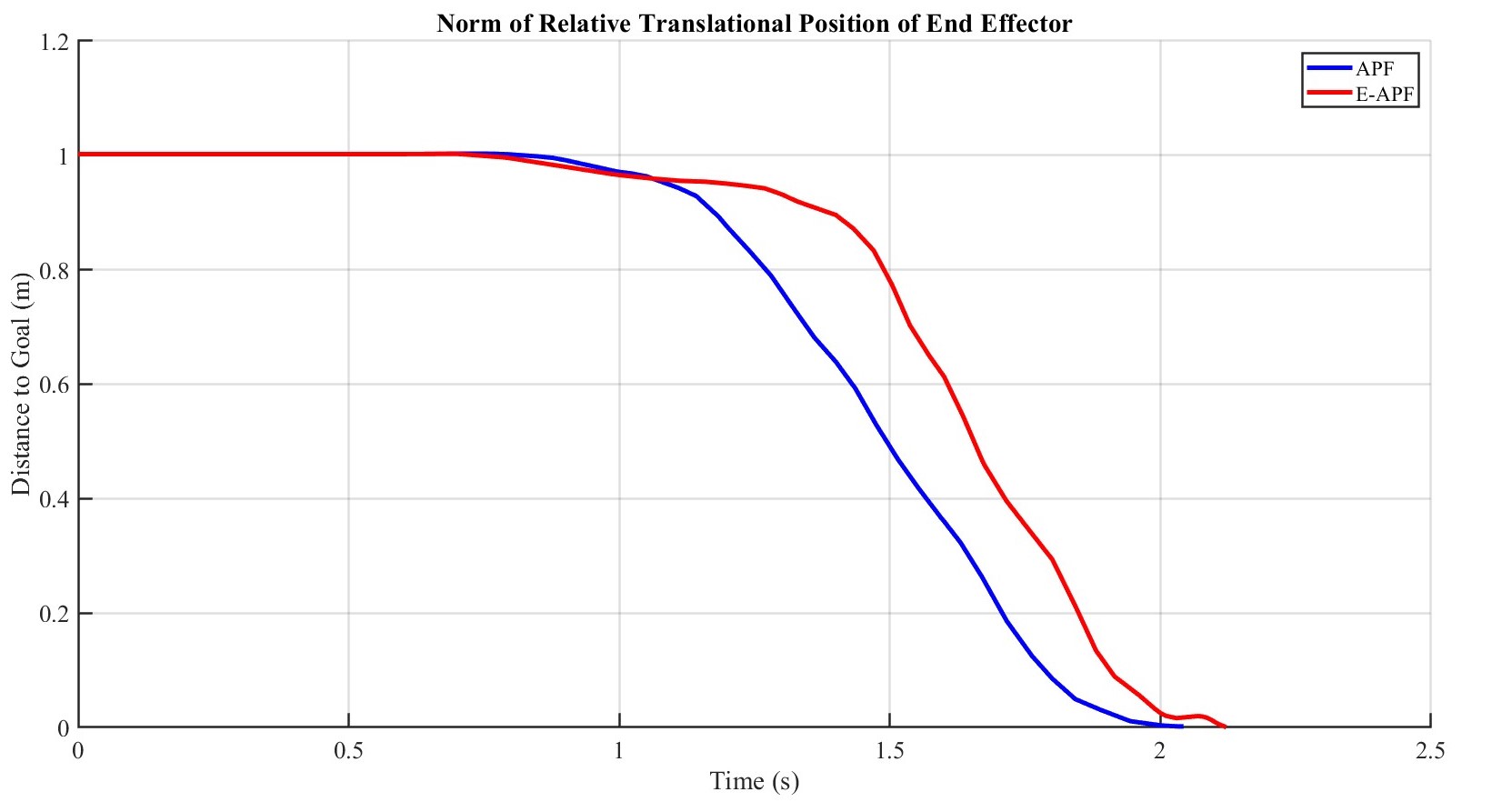}
    \caption{Norm of Relative Position of End Effector with respect to Goal}
    \label{fig:norm_rel_pos_goal}
\end{figure}

The joint angle trajectories as shown in Fig. \ref{fig:eapf_angles} show that the joint angles evolve smoothly over the course of motion, with no sudden jumps or discontinuities across all seven joints. This behavior is a direct consequence of the hybrid trajectory optimization framework that minimizes jerk while maintaining time efficiency. The trajectories exhibit optimal joint utilization where each joint contributes appropriately to the overall end-effector motion, with Joints 3, 5 and 6 showing moderate positive angular displacements that complement the primary motion generated by the other joints.
The plot of the relative position of the end effector with respect to the goal as shown in Fig. \ref{fig:norm_rel_pos_goal} depicts that using traditional APF the relative distance to the goal decreases gradually over time but shows a slower convergence rate, whereas the relative distance to the goal decreases more rapidly compared to APF, indicating faster convergence rate to the goal. This enhanced convergence rate is achieved through the velocity-dependent potential formulation, which effectively prevents the force equilibrium conditions as seen in traditional APF. E-APF might be taking a more cautious or indirect path to the goal initially due to a trade-off for better obstacle avoidance or smoother motion.

\begin{figure}[H]
    \centering
    \includegraphics[width=\linewidth]{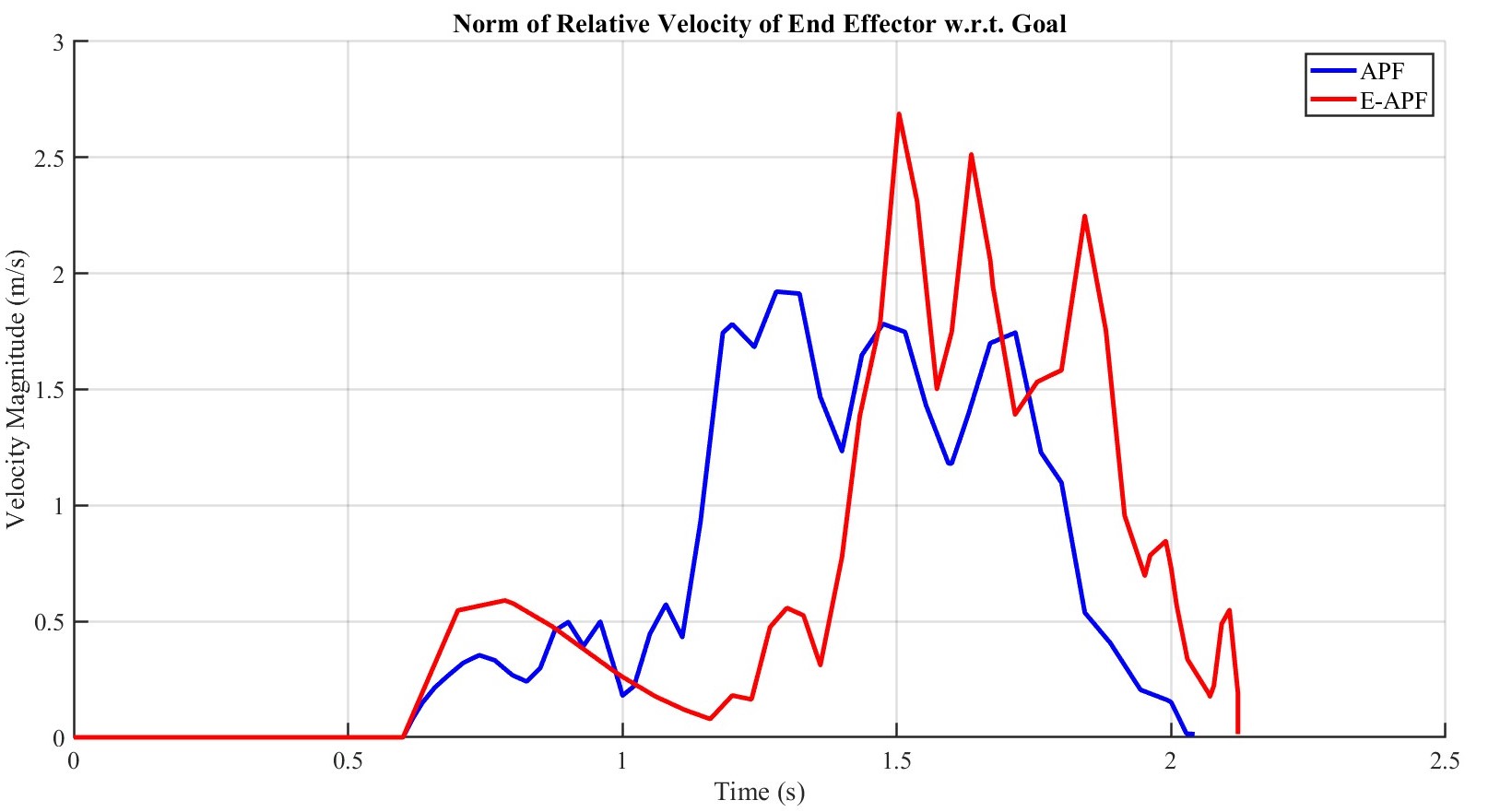}
    \caption{Norm of Relative Velocity of End Effector with respect to Goal}
    \label{fig:norm_rel_vel_goal}
\end{figure}

\begin{figure}[H]
    \centering
    \includegraphics[width=\linewidth]{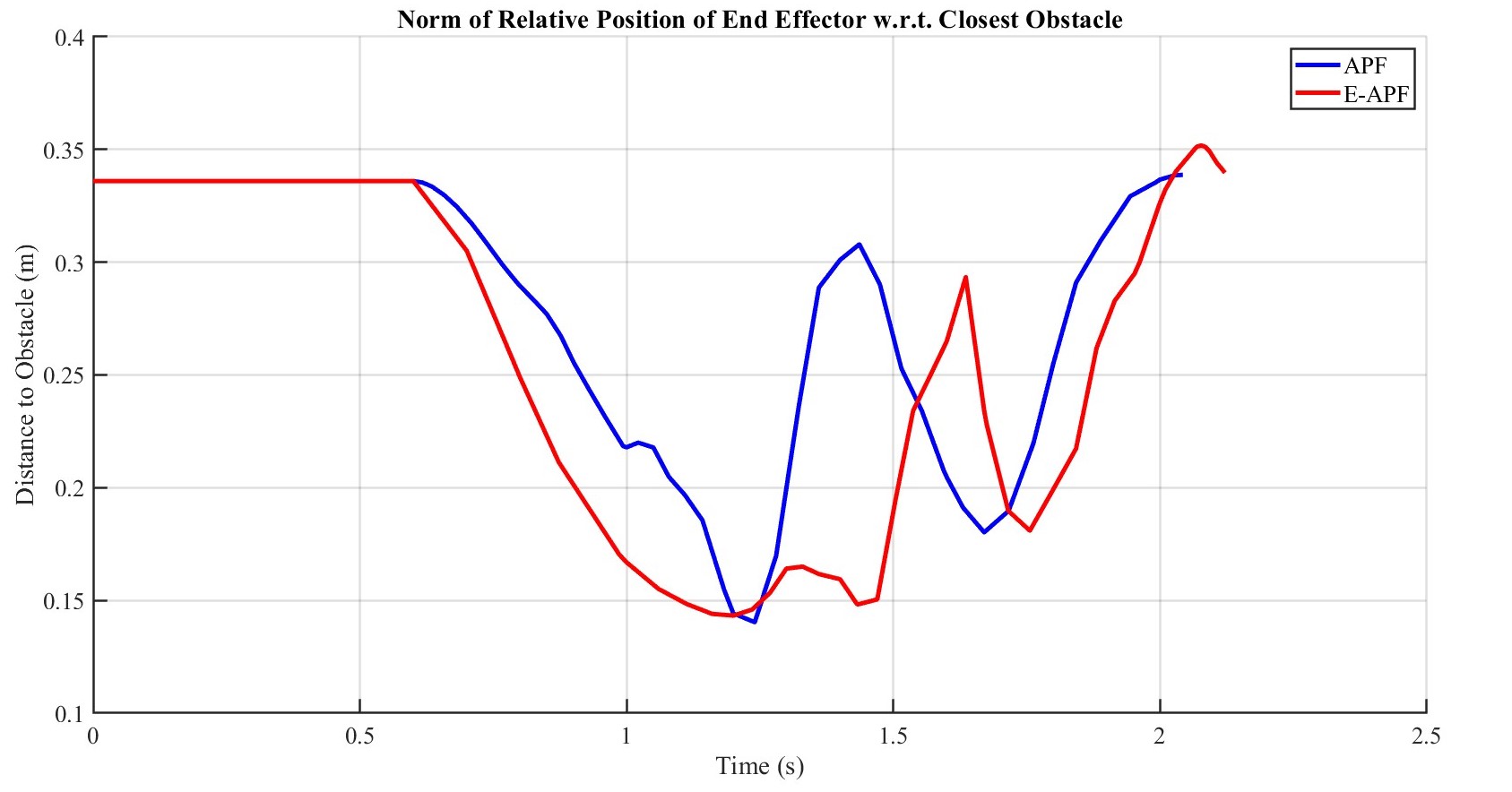}
    \caption{Norm of Relative Position of End Effector with respect to Obstacle}
    \label{fig:norm_rel_pos_obs}
\end{figure}

The relative velocity profile of the end effector with respect to the goal, shown in Fig. \ref{fig:norm_rel_vel_goal}, highlights a clear distinction between the two approaches. APF exhibits several peaks, indicating rapid accelerations and decelerations. E-APF, on the other hand, shows a more sustained and generally higher peak velocity around $t=1.5\,s$, and $t=1.75\,s$ during the later half of the trajectory. This is attributed to the obstacle avoidance strategy, where it needs to accelerate and decelerate more dynamically to navigate around obstacles while still progressing towards the goal. Moreover, E-APF's velocity profile appears to be more spread out in terms of magnitude. In contrast, the spiky yet low-magnitude nature of APF's velocity magnitude potentially results in jerky motion.
The relative position of end effector with respect to the closest obstacle as depicted in Fig. \ref{fig:norm_rel_pos_obs} shows that E-APF generally maintains a larger distance from the closest obstacle compared to APF throughout the trajectory, especially between $t=0.75\,s$ and $t=1.25\,s$, and again around $t=1.75\,s$. E-APF's smoother distance variations near the obstacle indicate better predictive obstacle avoidance behavior as compared to sudden directional change near the obstacle, as seen around $t=1.25\,s$ and $t=1.75\,s$. The pronounced dips and recoveries in E-APF's profile suggest that it actively adjusts its path to avoid close encounters, even if it means temporary deviations around $t=1.4\,s$. The increase in minimum distance from the obstacle for E-APF is a strong indicator of improved safety and robustness.

\begin{figure}[H]
    \centering
    \includegraphics[width=\linewidth]{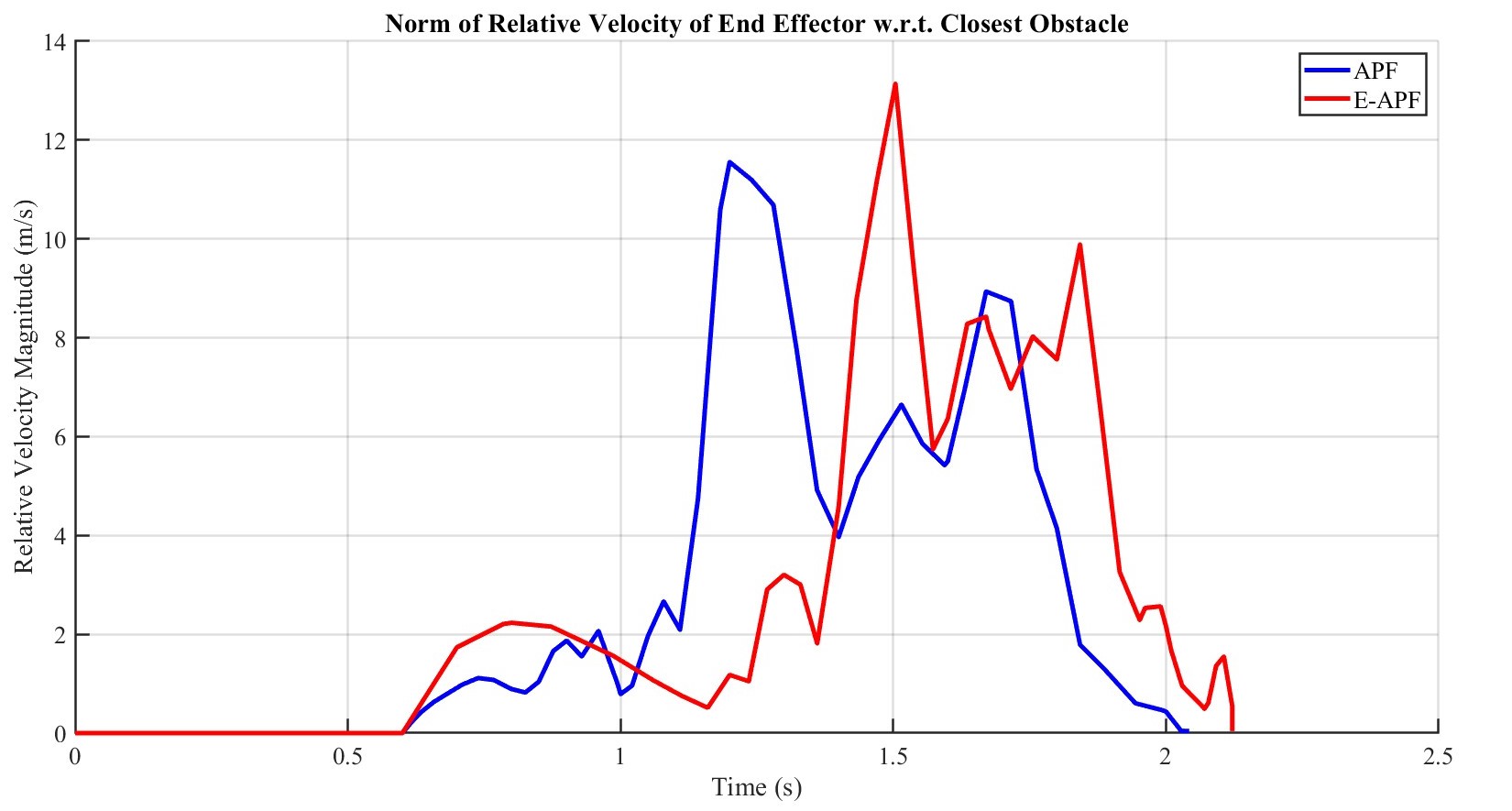}
    \caption{Norm of Relative Velocity of End Effector with respect to Obstacle}
    \label{fig:norm_rel_vel_obs}
\end{figure}

The relative velocity profile of the end effector with respect to the closest obstacle as shown in Fig. \ref{fig:norm_rel_vel_obs} shows that APF exhibits very high and sharp peaks in relative velocity, particularly around $t=1.25\,s$, exceeding $11\,m/s$. E-APF also has peaks at $t=1.5\,s$ (exceeding $12 m/s$) and another around $t=1.75\,s$. The extremely high peaks in APF's relative obstacle velocity could indicate rapid changes in direction or speed to avoid obstacles, which leads to instability or jerky movements. But in case of E-APF when the end effector is closest to the obstacle at around $t=1.25\,s$, the relative velocity with respect to the obstacle is quite low as compared to APF with peak at that time. The peaks of relative velocity profile in E-APF after $t=1.5\,s$ is strategically employed to ensure a safe distance from the obstacle.

\section{Conclusion}
In this paper, we have presented a novel Energy-based Artificial Potential Field (E-APF) for 7-DOF Kinova Gen3 robotic manipulator trajectory planning in dynamic, cluttered environments. By incorporating both position and velocity into the potential function, the proposed method effectively eliminates the local minima issue inherent in the traditional APF approach. The integration with a hybrid trajectory optimization scheme further ensures that generated paths are smooth, time-efficient, and dynamically feasible.  Future work includes extending this framework to multi-robot coordination and real-time adaptation in dynamic environments with moving obstacles.

\end{document}